\documentclass{article}
\usepackage{spconf,amsmath,graphicx}

\usepackage{graphicx,amsfonts,amsmath,amssymb,xcolor,ntheorem}
\usepackage{algorithm,algorithmic}

\usepackage{multirow}
\usepackage[noadjust]{cite}
\def\comment#1{}

\newcommand{\Amat}{{\bf A}}

\newcommand{\Dmat}{{\bf D}}

\newcommand{\Nmat}[0]{{{\bf N}}}

\newcommand{\Smat}[0]{{{\bf S}}}

\newcommand{\Xmat}{{\bf X}}
\newcommand{\Ymat}[0]{{{\bf Y}}}

\newcommand{\xv}{\boldsymbol{x}}
\newcommand{\yv}{\boldsymbol{y}}

\newcommand{\Sigmamat}{\boldsymbol{\Sigma}}

\newcommand{\muv}{\boldsymbol{\mu}}

\newcommand{\ie}{{\em i.e.}}

\def\LC{{$\text{L}^2\text{C}^2$}}
\title{Block-wise Lensless Compressive Camera}
\name{Xin Yuan\thanks{This work has been done in fall 2015 at Bell Labs.}, Gang Huang, Hong Jiang and Paul A. Wilford}
\address{Nokia Bell Labs, 600 Montain Avenue, Murray Hill, NJ, 07974, USA}
\begin{document}
%
\maketitle
\begin{abstract}
The existing lensless compressive camera ($\text{L}^2\text{C}^2$)~\cite{Huang13ICIP} suffers from low capture rates, resulting in low resolution images when acquired over a short time.	
In this work, we propose a new regime to mitigate these drawbacks.
We replace the global-based compressive sensing used in the existing $\text{L}^2\text{C}^2$ by the local block (patch) based compressive sensing. We use a single sensor for each block, rather than for the entire image, thus forming a multiple but spatially parallel sensor $\text{L}^2\text{C}^2$.
This new camera retains the advantages of existing $\text{L}^2\text{C}^2$ while leading to the following additional benefits:
1) Since each block can be very small, {\em e.g.}$~8\times 8$ pixels, we only need to capture $\sim 10$ measurements to achieve reasonable reconstruction. Therefore the capture time can be reduced significantly.
2) The coding patterns used in each block can be the same, therefore the sensing matrix is only of the block size compared to the entire image size in existing $\text{L}^2\text{C}^2$. This saves the memory requirement of the sensing matrix as well as speeds up the reconstruction.
3) Patch based image reconstruction is fast and since real time stitching algorithms exist, we can perform real time reconstruction.
4) These small blocks can be integrated to any desirable number, leading to ultra high resolution images while retaining fast capture rate and fast reconstruction.
We develop multiple geometries of this block-wise $\text{L}^2\text{C}^2$ in this paper.
We have built prototypes of the proposed  block-wise $\text{L}^2\text{C}^2$ and demonstrated excellent results of real data.
\end{abstract}
\vspace{-4mm}
\section{Motivation}
\label{Sec:Mo}
\vspace{-2mm}
Inspired by compressive sensing (CS)~\cite{Donoho06ITT,Candes06ITT}, diverse compressive cameras~\cite{Duarte08SPM,Wagadarikar08CASSI,Patrick13OE,Reddy11CVPR,Hitomi11ICCV,Yuan14CVPR,Tsai15OE,Tsai15OL,Sun16OE,Cao16SPM} have been built.
The single-pixel camera~\cite{Duarte08SPM}, which implemented the compressive imaging in space, is an elegant architecture to prove the concept of CS.
However, the hardware elements used in the single-pixel camera are more expensive than conventional CCD or CMOS cameras, thus limit the applications of this new imaging regime.
The lensless compressive camera ($\text{L}^2\text{C}^2$), proposed in~\cite{Huang13ICIP}, enjoys low-cost property and has demonstrated excellent results using advanced reconstruction algorithms~\cite{Yuan16SJ,Yuan15Lensless,Yuan15GMM}.

\begin{figure}[htbp!]
	\centering
	\includegraphics[width=.5\textwidth]{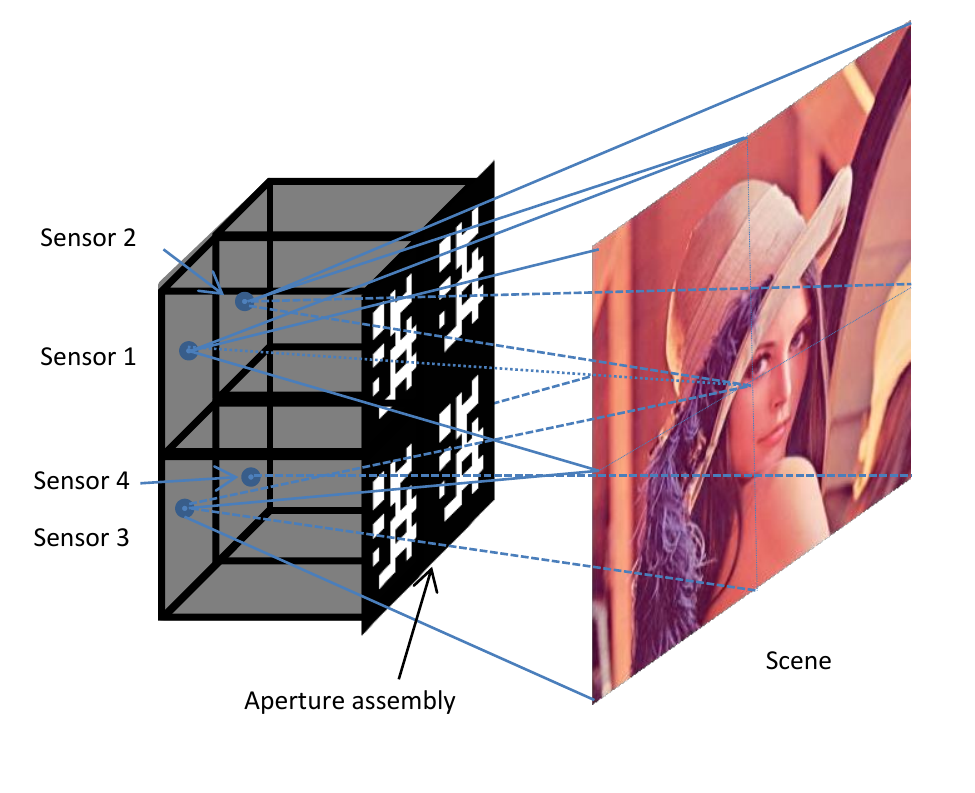}\\
	\caption{{\small Geometry and imaging process of the block-wise lensless compressive camera. Four sensors are shown in this example.
			Each sensor will capture a fraction of the scene. 
			These fractions can be overlapping. The image is reconstructed via first performing patch-based inversion (reconstruction) and then stitching these patches.} }
	\label{fig:ge}
\end{figure}

Though enjoying various advantages, the existing $\text{L}^2\text{C}^2$~\cite{Huang13ICIP} suffers from low capture rates.
Specifically, the currently used sensor can only capture the scene at around $10$Hz.
For a $64\times 64$ image, if we desire a high resolution image, we need on the order of $10\%$ measurements (relative to the pixel numbers). This requires about 1 minute with current aperture switching technique. 
This is far from our real time requirement as cameras are built to get instant images/videos.
In order to obtain images in a shorter time, we have to sacrifice the spatial resolution, \ie~providing a low-resolution image.
An alternative solution is to increase the refresh rate of the aperture assembly (Fig.~\ref{fig:ge}).
However, even we can achieve a higher refresh rate using an expensive hardware, the sensor needs a relative long time to integrate the light or a more expensive sensor is required.

There are strong advantages with the \LC, directly implementing compressive sensing. These include minimizing problems of lenses sensor. But with current approaches, speed is an issue.
In this paper, we propose the block-wise \LC to mitigate this problem.
\section{Block-wise Lensless Compressive Camera}
\begin{figure}[htbp!]
	\centering
	\includegraphics[width=.5\textwidth]{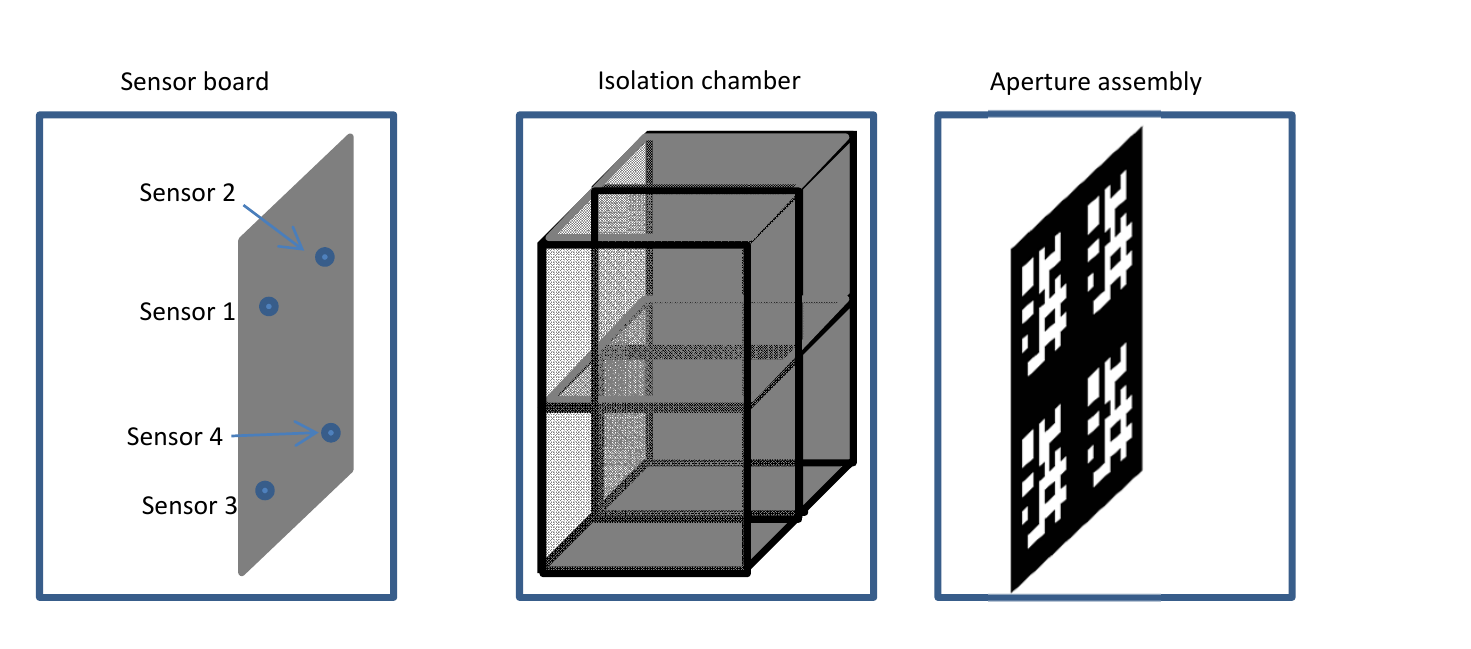}\\
	\vspace{-5mm}
	\caption{{\small Components of the block-wise lensless camera. Each parts can be obtained with off-the-shelf components.}}
	\label{fig:part}
\end{figure}

Figure~\ref{fig:ge} depicts the geometry of the proposed block-wise \LC.
It consists of three components as shown in Fig.~\ref{fig:part}:
a) the sensor board which contains multiple sensors and each one corresponding to one block,
b) the isolation chamber which prevents the light from other blocks,
and
c) the aperture assembly which can be the same as that used in the \LC~\cite{Huang13ICIP}.

Note that the block sizes can be different such that each block can reconstruct different resolution image fractions, thus leading to multi-scale compressive camera~\cite{BradyNature12}.
The pattern used for each block can also be different and can be adapted to the content of the image part, thus leading to adaptive compressive sensing~\cite{Yuan13ICIP}.
In this work, we consider each block uses the same pattern as this will not only save the memory to store these patterns but also enables fast reconstruction for each block.  

%

This new block-wise \LC enjoys the following advantages compared to the existing \LC.
\begin{itemize}
	\setlength\itemsep{-0.4em}
	\vspace{-1mm}
	\item[$i)$] Since each block can be very small, {\em e.g.} $8\times 8$ pixels, we only need to capture a small number of measurements to achieve high resolution reconstruction. Therefore the capture time can be short.
	\item[$ii)$] The coding patterns used in each block can be the same, therefore the sensing matrix is only of the block size. This saves the memory requirement of sensing matrix as well as speeds up the reconstruction.
	\item[$iii)$] Patch based image reconstruction is fast and since real time stitching algorithms exist, we can perform real time reconstruction.
	\item[$iv)$] These blocks can be integrated to any desirable number, leading to extra high resolution images~\cite{BradyNature12} while retaining the capture rate and fast reconstruction;
	\item[$v)$]  The sensor layer can be very close to the aperture assembly, which leads to small size camera. In particular, the thickness of the camera will be extremely small~\cite{Asif_flatCam}.
\end{itemize}

\subsection{Overlapping Regions and Stitching \label{Sec:overlap_Far}}
From the simple geometry shown in Fig.~\ref{fig:ge}, we can see that if the scene is far from the sensor, we will have significant overlapping regions. 
This will be discussed in next section.
On the other hand, the image stitching algorithms are usually based on the features within the overlapping areas to perform registration.
Since the angular resolution is limited by the distance between sensors and the aperture assembly, we can adapt this distance in different applications.

\begin{figure}[htbp!]
	\centering
	\includegraphics[width=.5\textwidth]{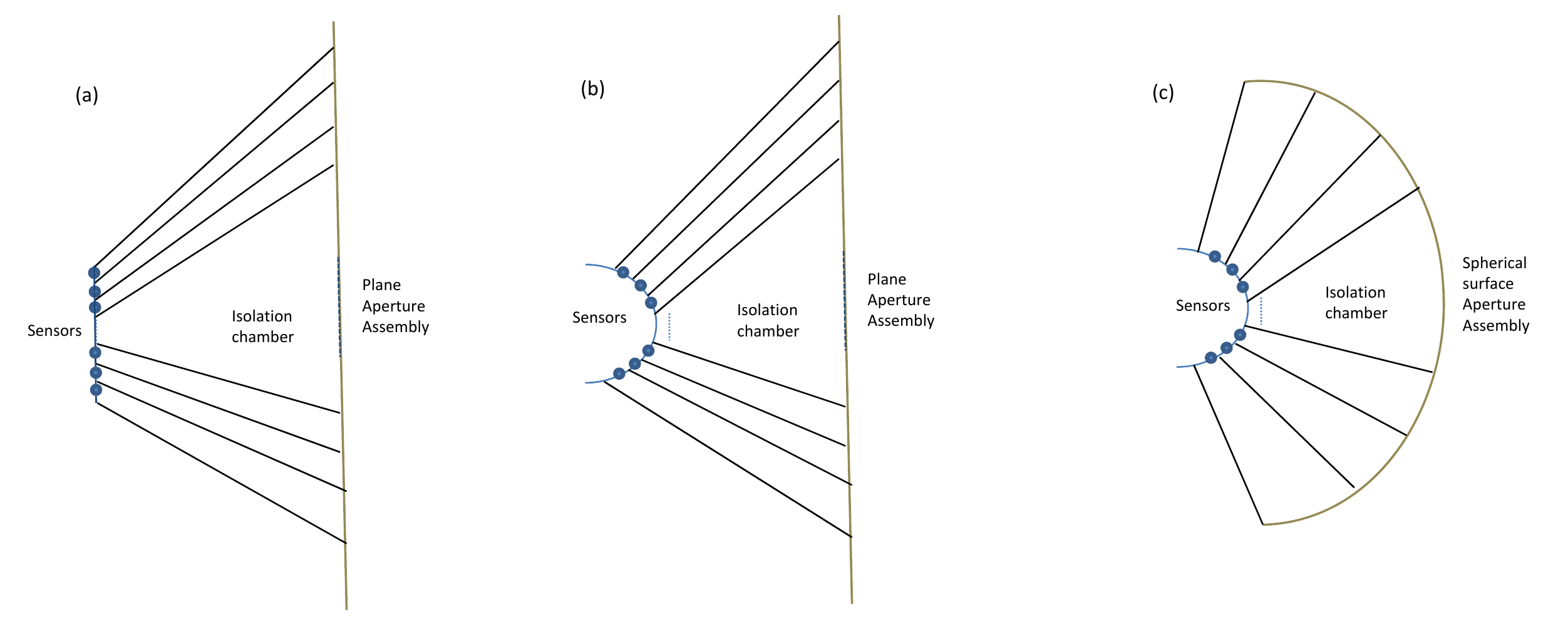}\\
	\vspace{-5mm}
	\caption{{\small Cross-sectional view of the ``Concentration-Sensor Regime", where the aperture assembly can be a plane (a-b) or a spherical surface (c). The sensors can be mounted on a plane (a) or a sphere (b-c).}}
	\label{fig:c_sensor1}
\end{figure}
\begin{figure}[htbp!]
	\centering
	\vspace{-5mm}
	\includegraphics[width=.5\textwidth]{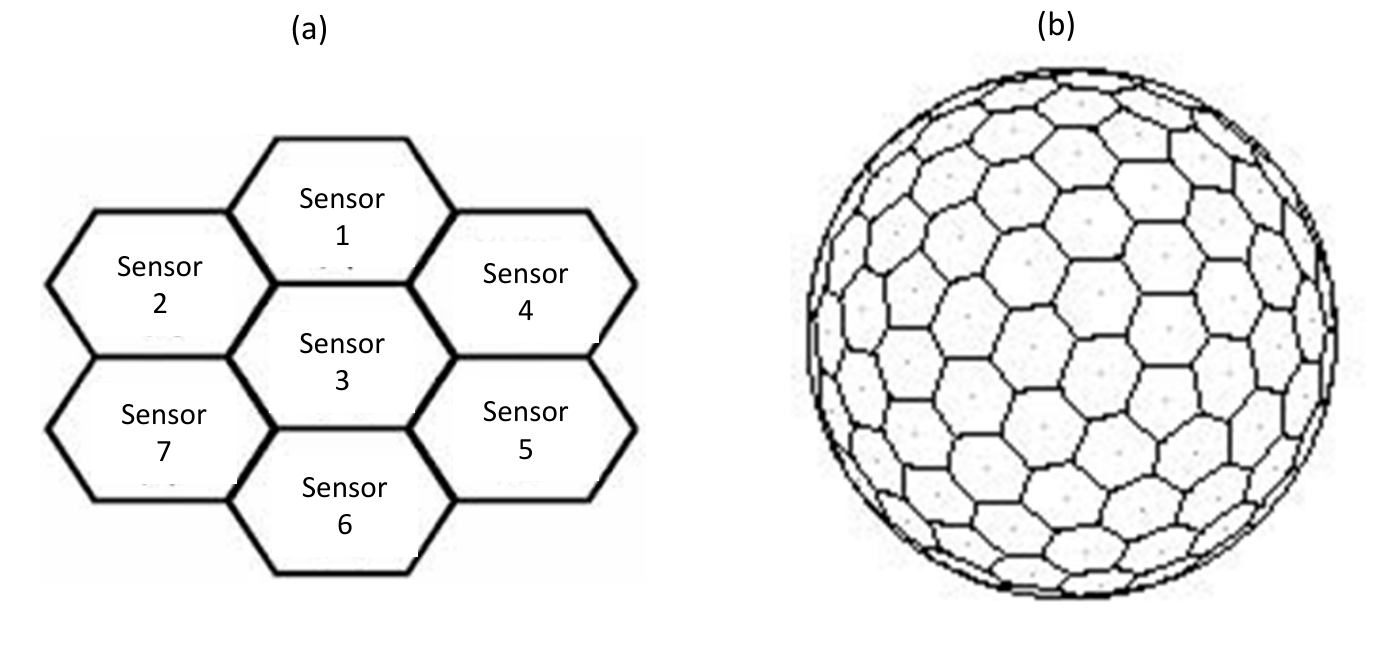}\\
	\vspace{-5mm}
	\caption{{\small Sensor layout of the ``Concentration-Sensor Regime". Each sensor covers a hexagon area (a) and the senor array forms a sphere (b).}}
	\label{fig:c_sensor2}
\end{figure}
\begin{figure}[htbp!]
	\centering
	\includegraphics[width=.5\textwidth]{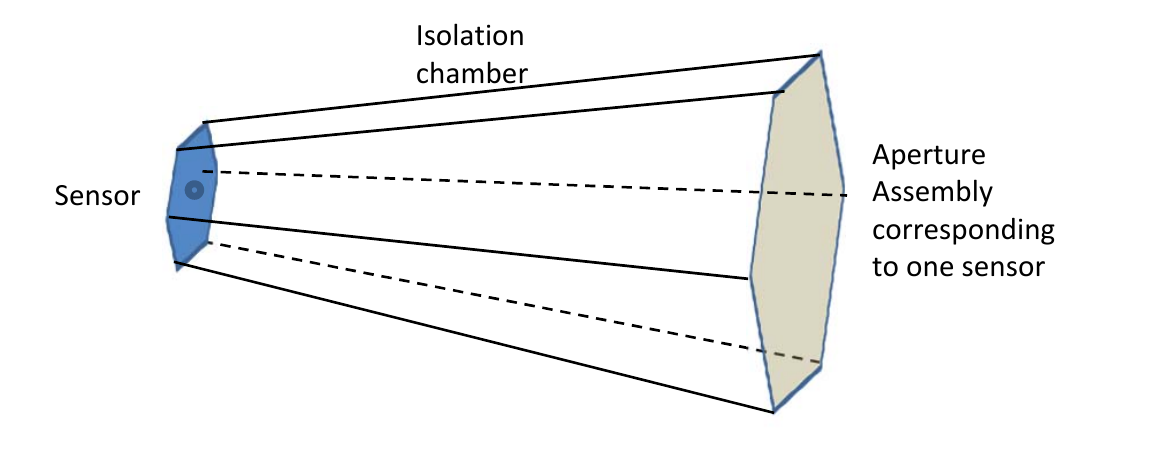}\\
	\vspace{-5mm}
	\caption{{\small Isolation chamber in the ``Concentration-Sensor Regime". }}
	\label{fig:box}
\end{figure}
\subsection{Concentration-Sensor Regime}
In order to mitigate the problem mentioned in Section~\ref{Sec:overlap_Far}, \ie~the low angular resolution for far scenes, we propose the following {\bf Concentration-Sensor Regime}, where the sensors are put together in a ``cellular" shape. The aperture assembly can be a plane or a spherical surface as shown in Fig.~\ref{fig:c_sensor1}.
The planer aperture assembly in Fig.~\ref{fig:c_sensor1}(a-b) can be replaced by a spherical aperture as shown in Fig.~\ref{fig:c_sensor1}(c).
The sensor layout of the spherical regime is detailed in Fig.~\ref{fig:c_sensor2}.
In this configuration, the isolation chamber will be a ``trumpet" shape starting from the sensor to the aperture assembly, as demonstrated in Fig.~\ref{fig:box}. Note the configuration in Fig.~\ref{fig:c_sensor1} (c) and Fig.~\ref{fig:c_sensor2} is a wide angle camera.


\section{Fast Reconstruction}
We consider that the patterns used for each block are the same. The measurements can be expressed as
\begin{eqnarray}\label{eq:mea}
\Ymat &=& \Amat \Xmat + \Nmat,
\end{eqnarray}
where $\Xmat \in {\mathbb R}^{P\times N_p}$ with $P$ denoting the dimension of the vectorized block (with $\sqrt{P}\times \sqrt{P}$ pixels) and $N_P$ is the number of blocks used in the camera. $\Ymat\in{\mathbb R}^{M \times P}$ with $M\ll P$ denotes the measurements and each column corresponds to one block (measurements captured by one sensor), and $\Nmat$ signifies the measurement noise.
$\Amat \in {\mathbb R}^{M\times N_p}$ is the sensing matrix, which can be random~\cite{cs_Candes06randomProj} or the Hadamard matrix with random permutation~\cite{Yuan15Lensless}.

\subsection{Dictionary Based Inversion}
Introducing a basis or (block-based) dictionary $\Dmat$, equation (\ref{eq:mea}) can be reformulated as
\begin{eqnarray}\label{eq:Dmea}
\Ymat &=& \Amat \Dmat \Smat + \Nmat,
\end{eqnarray} 
where $\Dmat \in {\mathbb R}^{P\times Q}$ can be an orthonormal basis with $Q=P$ or an over-complete dictionary~\cite{Aharon06TSP} with $P\ll Q$. This dictionary can be pre-learned for fast inversion or learned {\em in situ}~\cite{Yuan15JSTSP,Yuan_16_OE}.
$\Smat \in{\mathbb R}^{Q\times N_p}$ is desired to be sparse so that various $\ell_1$-based algorithms can be used to solve the following problem~\cite{Liao14GAP,Yuan16ICIP_GAP}
\begin{eqnarray} \label{eq:L1}
\min \|\Smat\|_1, &{\text {~~subject to~~}}& \Ymat = \Amat \Dmat \Smat,
\end{eqnarray}
given $\Amat, \Ymat$ and $\Dmat$ or variant problems~\cite{Beck09IST}.

Diverse algorithms have been proposed to solve the above problem and we will use the advanced Gaussian mixture model described below since it does not need any iteration as closed-form analytic solution exists.

\subsection{Closed-form Inversion via Gaussian Mixture Models \label{Sec:GMM}}
The Gaussian mixture model (GMM) has recently been re-recognized as an efficient dictionary learning algorithm~\cite{Chen10SPT,Yu12IPT,Yang14GMM,Yang14GMMonline,Yuan15GMM}.
Recall the image blocks  $\Xmat \in{\mathbb R}^{P \times N_p}$ extracted from the image. For $i$-th patch $\xv_i$, it is modeled by a GMM with $K$ Gaussian components~\cite{Chen10SPT}:
\begin{equation}\label{eq:xiGMM}
\xv_i \sim \sum_{k=1}^K \pi_k {\cal N}(\muv_k, \Sigmamat_k),
\end{equation} 
where $\{\muv_k, \Sigmamat_k\}_{k=1}^K$ represent the mean and covariance matrix of $k$-th Gaussian, and $\{\pi_k\}_{k=1}^K$ denote the weights of these Gaussian components, and $\sum_k \pi_k = 1$. 

Dropping the block index $i$, in a linear model $\yv = \Amat\xv+ {\boldsymbol\epsilon}$, ${\boldsymbol\epsilon} \in {\cal N}(\mathbf{0},{\bf R})$, if $\xv\sim p({\xv})$ in (\ref{eq:xiGMM}), then $p({\xv}|{\yv})$ has the following analytical form~\cite{Chen10SPT}
\begin{align}\label{eq_GMMinv}
p(\xv|\yv) = \sum_{k=1}^K \tilde{\pi}_k \mathcal{N}(\xv|\tilde{{\boldsymbol \mu}}_k,\tilde{{\bf \Sigma}}_k)
\end{align}
where
\begin{align}\label{eq:xv_pdf}
\tilde{\pi}_k &= \frac{\pi_k {\cal N}({\bf y}|\Amat \xv_k, {\bf R}^{-1} + \Amat  {\bf \Sigma}_k \Amat ^T)}{\sum_{l=1}^K \pi_l {\cal N}({\bf y}|\Amat  {\bf x}_l, {\bf R}^{-1} + \Amat {\bf \Sigma}_l \Amat ^T)}\\
\tilde{{\bf\Sigma}}_k &= (\Amat ^T {\bf R} \Amat  + {\bf \Sigma}_k^{-1})^{-1},\\
\tilde{\boldsymbol \mu}_k &= \tilde{{\bf \Sigma}}_k(\Amat ^T {\bf R} {\bf y} + {\bf \Sigma}_k^{-1}{\boldsymbol \mu}_k).
\end{align}
While (\ref{eq_GMMinv}) provides a posterior distribution for ${\xv}$, we obtain the point estimate of $\hat{\xv}$ via the posterior mean:
\begin{eqnarray}\label{eq:txhatmean}
{\mathbb E}[\hat{\xv}]&=& \sum_{k=1}^K \tilde{\pi}_k \tilde{\boldsymbol \mu}_k.
\end{eqnarray}
which is a closed-form solution.

Note that $\{\pi_k, \muv_k, \Sigmamat_k\}_{k=1}^K$ are pre-trained on other datasets and given $\Amat$, $\tilde{{\bf\Sigma}}_k$ only needs to be computed once and saved. Same techniques can be used for $\Amat\Sigmamat_k\Amat^T$.
The only computation left for each block is to calculate $\{\tilde{\boldsymbol \mu}_k, \tilde{\pi}_k\}$, which can be obtained very efficiently.

Most importantly, no iteration is required using the above GMM method, leading to {\em real-time} reconstruction for each block.
Furthermore, each block can be reconstructed in parallel with GPU. 
%
Since real time stitching algorithm exists, after blocks are obtained via the GMM, we can get the entire image instantly. Alternatively, we can pre-train a neural network for the inversion and obtain the reconstruction in real time~\cite{Kulkarni2016CVPR}.

\begin{figure}[h!]
	\centering
	\vspace{-5mm}
	\includegraphics[width=.5\textwidth]{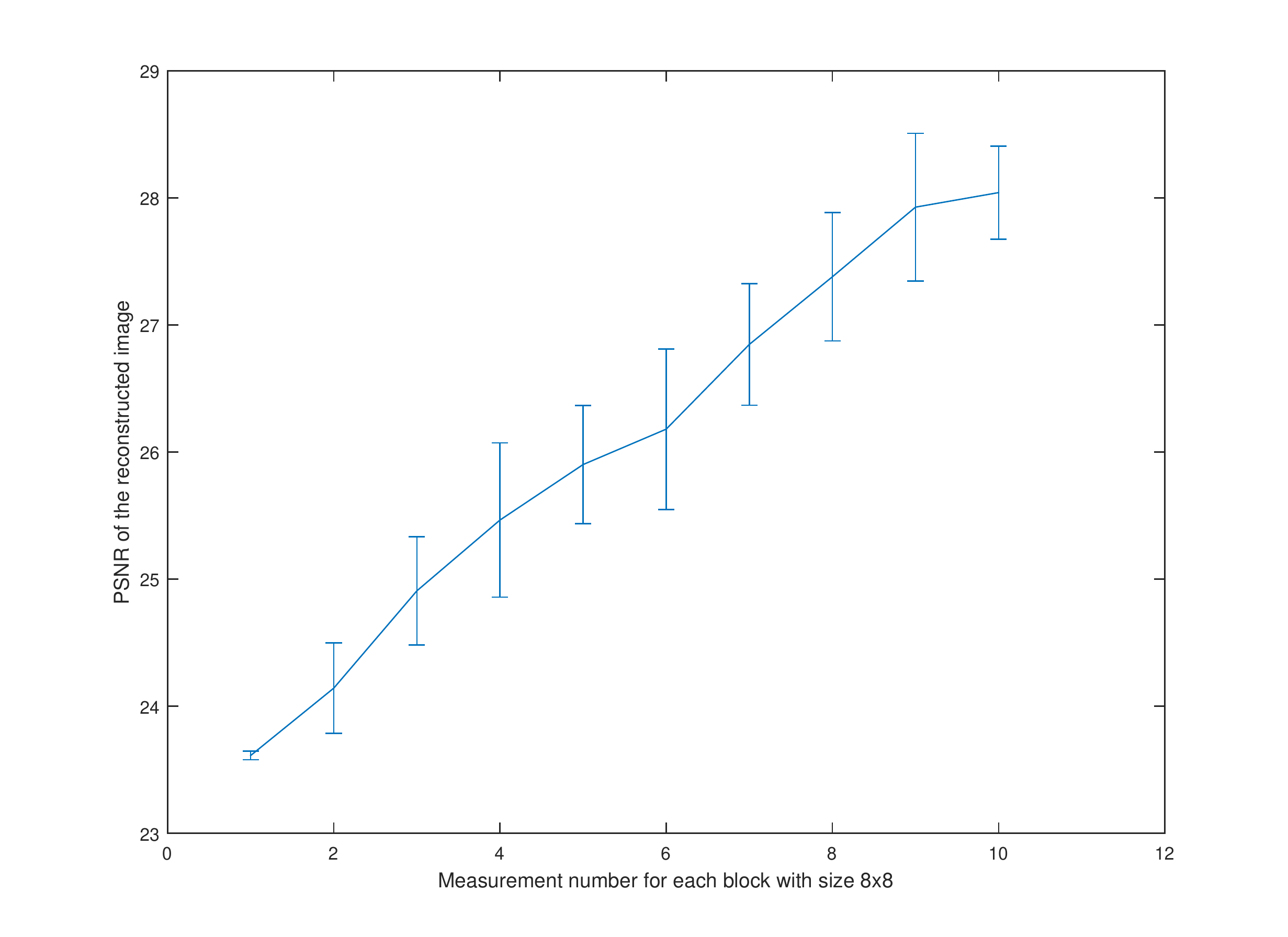}\\
	\vspace{-5mm}
	\caption{{\small PSNR of the reconstructed image using different number of measurements for each $8\times 8$ (pixels) block. 10 trials were performed with different random binary coding patterns.}}
	\label{fig:psnr}
\end{figure}
\begin{figure}[h!]
	\centering
	\includegraphics[width=.5\textwidth]{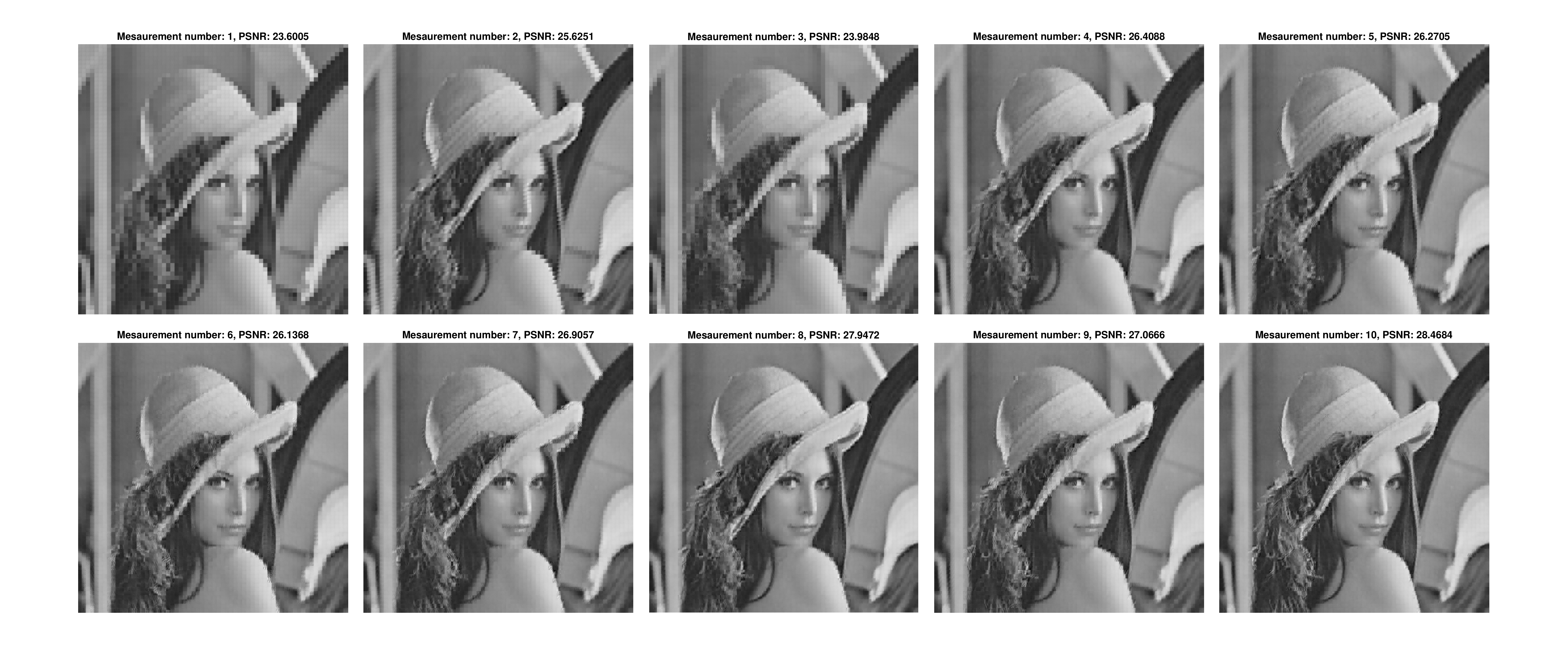}\\
	\vspace{-5mm}
	\caption{{\small Example of reconstructed images using different number of measurements (from top-left to bottom-right: 1 to 10) for each $8\times 8$ pixels block.}}
	\label{fig:imge}
\end{figure}

\section{Results}
In this section, we first conduct the simulation to verify the proposed approach and then describe the hardware prototypes we have built to demonstrate real data results.
\subsection{Simulation}
We consider each block with size $8\times 8$ pixels and reconstruct the images with different numbers of measurements.
The image is of size $512\times 512$.
We assume that there is no overlapping among blocks and the image part is ideally captured by different sensors.
The PSNR (Peak-Signal-to-Noise-Ratio) of the reconstructed image is used as a metric and plotted in Fig.~\ref{fig:psnr} with exemplar reconstructed images shown in Fig.~\ref{fig:imge}. The GMM is used for reconstruction. We can observe that even a single measurement in each block provides excellent reconstruction. The running time for reconstruction is $<0.01$ seconds using an i7 CPU. 

\begin{figure}[h!]
	\centering
	\includegraphics[width=.5\textwidth]{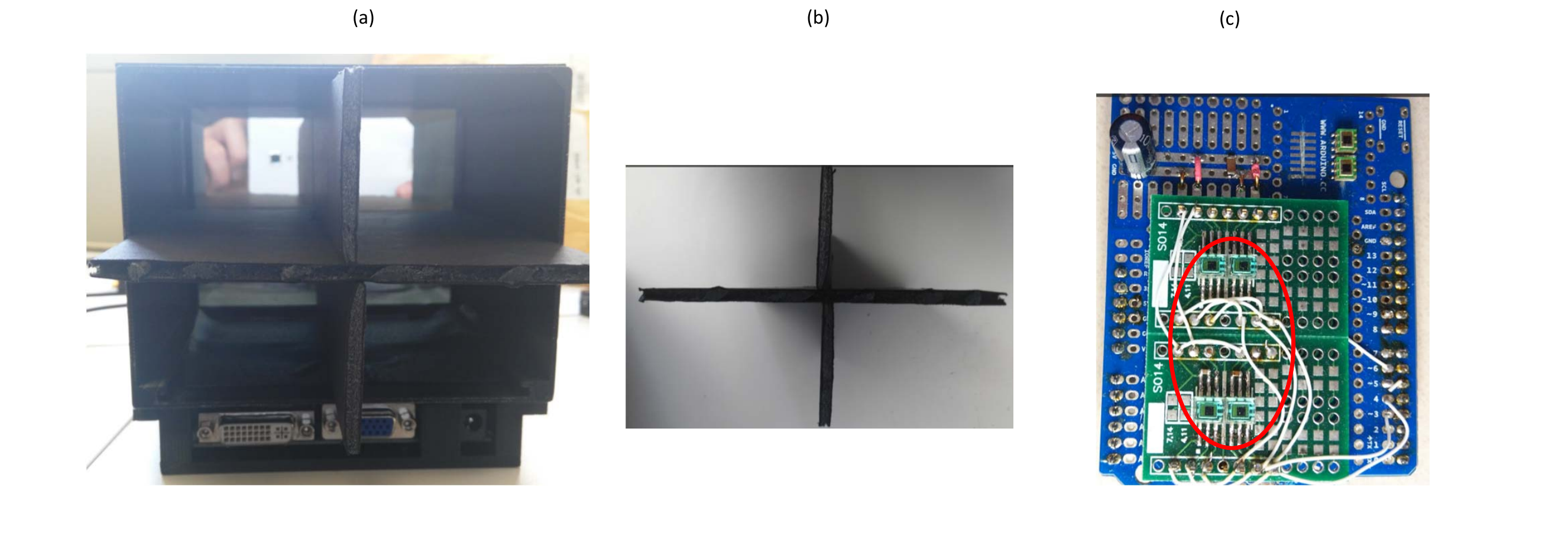}\\
	\vspace{-5mm}
	\caption{{\small Prototype with 4 sensors used to prove the concept, (a) LCD used as aperture assembly with the isolation chamber (b) and (c) the 4 sensors.}}
	\label{fig:block_camera}
	\vspace{-5mm}
\end{figure}
\begin{figure}[h!]
	\centering
	\includegraphics[width=.45\textwidth]{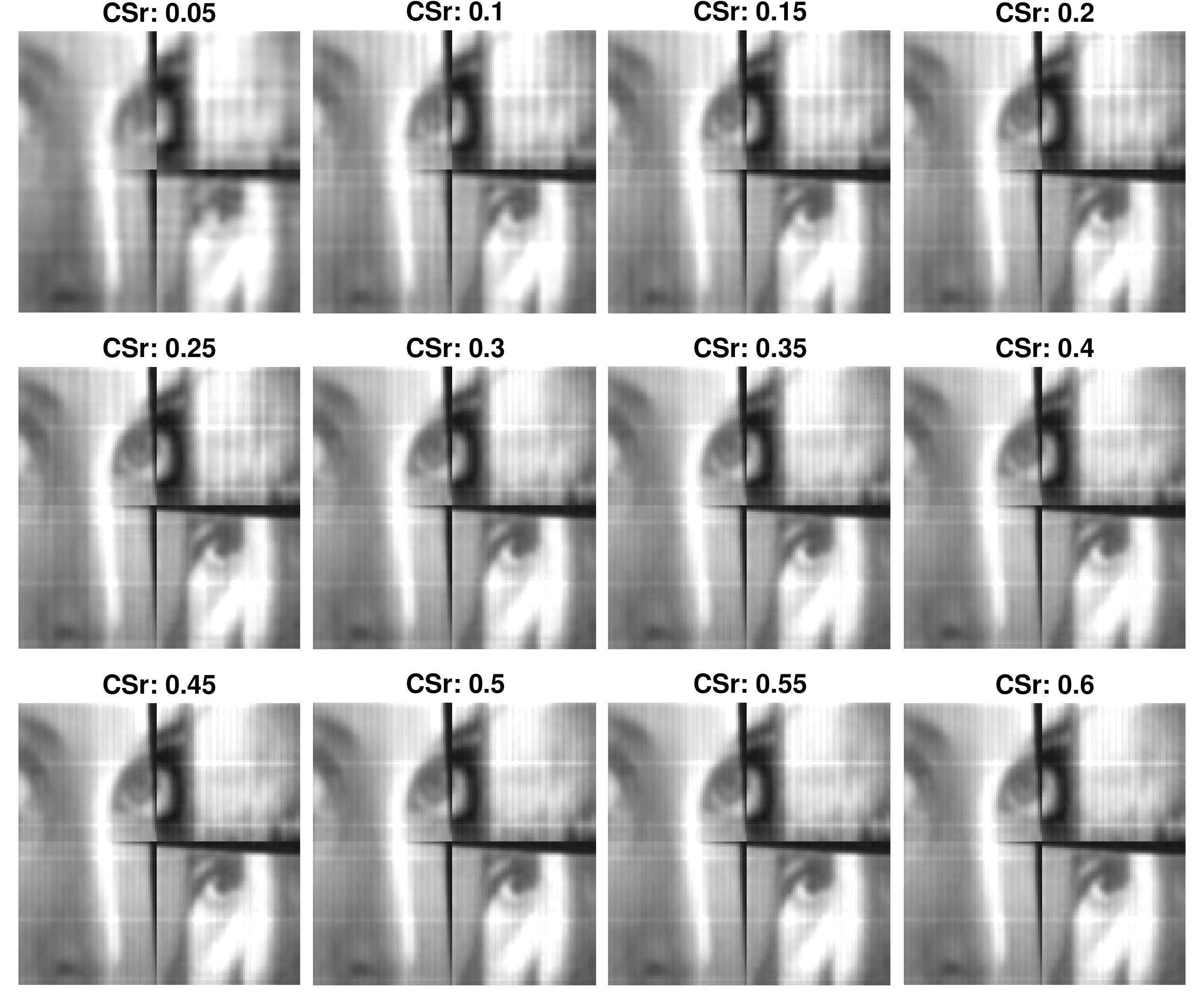}\\
	\vspace{-3mm}
	\caption{{\small Results from the measurements taken by the prototype with 4 sensors.}}
	\label{fig:block_result2}
	\vspace{-5mm}
\end{figure}

\subsection{Proof of Concept Prototypes and Results}
We first used 4 sensors to build the block-wise \LC to prove the concept, with photos shown in Fig.~\ref{fig:block_camera}.
We used $32\times 32$ pixels pattern for each sensor and these patterns are shared across these 4 sensors. The scene is a photo printed on a paper and pasted on a wall.
Without employing any stitching algorithm, one result is shown in Fig.~\ref{fig:block_result2}, where the CSr is defined as the number of measurements relative to pixels in the reconstructed image.
It can be observed that good results can be obtained by only using $10\%$ of the pixel data (CSr = 0.1), which is 102 measurements captured by each sensor.

\begin{figure}[h!]
	\centering
	\includegraphics[width=.5\textwidth]{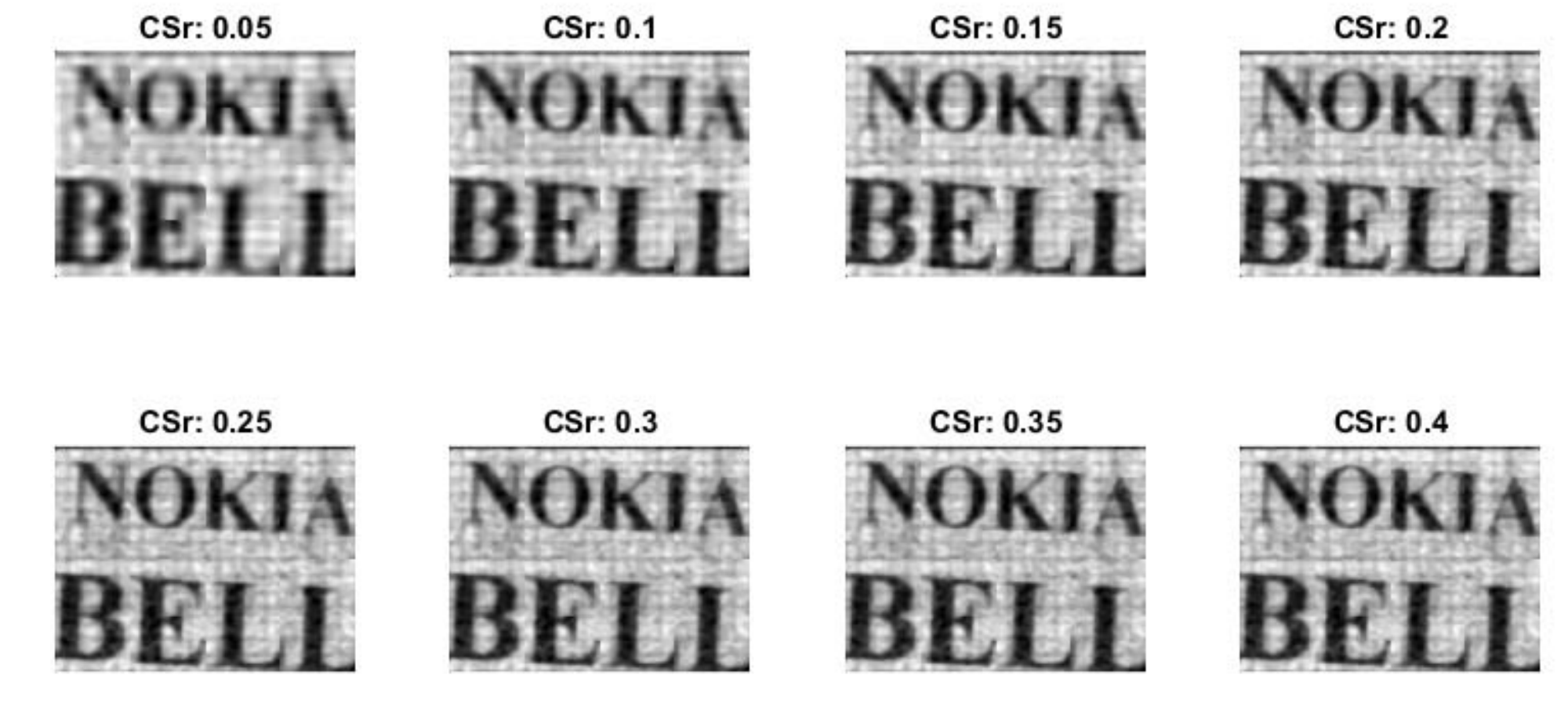}\\
	\vspace{-3mm}
	\caption{{\small Results from the measurements taken by the prototype with 16 sensors.}}
	\label{fig:block_xy}
\end{figure}

Next we show the reconstruction results with 16 sensors (a $4\times4$ array) in Fig.~\ref{fig:block_xy}, where we restricted the light to test the performance of the camera in dark environment. It can be observed that even when CSr$=0.05$ (as each block is of $16\times16$ pixels, CSr$=0.05$ denotes each sensor only captured 12 measurements), we can still get reasonable results. 
%

\vspace{-6mm}
\section{Conclusions}
%
We have proposed the block-wise lensless compressive camera to mitigate the speed issue existing in the current \LC. Multiple geometries of this block-wise $\text{L}^2\text{C}^2$ have been developed.
Prototypes have been built to demonstrate the feasibility of the proposed imaging architecture.
Excellent results of real data have been shown to verify the fast capture and real time reconstruction of the proposed camera.

\small
\bibliographystyle{IEEEbib}

\begin{thebibliography}{10}
	
	\bibitem{Huang13ICIP}
	G.~Huang, H.~Jiang, K.~Matthews, and P.~Wilford,
	\newblock ``Lensless imaging by compressive sensing,''
	\newblock {\em IEEE International Conference on Image Processing}, 2013.
	
	\bibitem{Donoho06ITT}
	D.~L. Donoho,
	\newblock ``Compressed sensing,''
	\newblock {\em IEEE Transactions on Information Theory}, vol. 52, no. 4, pp.
	1289--1306, April 2006.
	
	\bibitem{Candes06ITT}
	E.~J. Cand\`{e}s, J.~Romberg, and T.~Tao,
	\newblock ``Robust uncertainty principles: Exact signal reconstruction from
	highly incomplete frequency information,''
	\newblock {\em IEEE Transactions on Information Theory}, vol. 52, no. 2, pp.
	489--509, February 2006.
	
	\bibitem{Duarte08SPM}
	M.~F. Duarte, M.~A. Davenport, D.~Takhar, J.~N. Laska, T.~Sun, K.~F. Kelly, and
	R.~G. Baraniuk,
	\newblock ``Single-pixel imaging via compressive sampling,''
	\newblock {\em IEEE Signal Processing Magazine}, vol. 25, no. 2, pp. 83--91,
	2008.
	
	\bibitem{Wagadarikar08CASSI}
	A.~Wagadarikar, R.~John, R.~Willett, and D.~J. Brady,
	\newblock ``Single disperser design for coded aperture snapshot spectral
	imaging,''
	\newblock {\em Applied Optics}, vol. 47, no. 10, pp. B44--B51, 2008.
	
	\bibitem{Patrick13OE}
	P.~Llull, X.~Liao, X.~Yuan, J.~Yang, D.~Kittle, L.~Carin, G.~Sapiro, and D.~J.
	Brady,
	\newblock ``Coded aperture compressive temporal imaging,''
	\newblock {\em Optics Express}, pp. 698--706, 2013.
	
	\bibitem{Reddy11CVPR}
	D.~Reddy, A.~Veeraraghavan, and R.~Chellappa,
	\newblock ``{\rm P2C2}: Programmable pixel compressive camera for high speed
	imaging,''
	\newblock {\em IEEE Computer Vision and Pattern Recognition (CVPR)}, 2011.
	
	\bibitem{Hitomi11ICCV}
	Y.~Hitomi, J.~Gu, M.~Gupta, T.~Mitsunaga, and S.~K. Nayar,
	\newblock ``Video from a single coded exposure photograph using a learned
	over-complete dictionary,''
	\newblock in {\em IEEE International Conference on Computer Vision (ICCV)},
	2011.
	
	\bibitem{Yuan14CVPR}
	X.~Yuan, P.~Llull, X.~Liao, J.~Yang, G.~Sapiro, D.~J. Brady, and L.~Carin,
	\newblock ``Low-cost compressive sensing for color video and depth,''
	\newblock in {\em IEEE Conference on Computer Vision and Pattern Recognition
		(CVPR)}, 2014.
	
	\bibitem{Tsai15OE}
	T.-H. Tsai, X.~Yuan, and D.~J. Brady,
	\newblock ``Spatial light modulator based color polarization imaging,''
	\newblock {\em Optics Express}, vol. 23, no. 9, pp. 11912--11926, May 2015.
	
	\bibitem{Tsai15OL}
	T.-H. Tsai, P.~Llull, X.~Yuan, D.~J. Brady, and L.~Carin,
	\newblock ``Spectral-temporal compressive imaging,''
	\newblock {\em Optics Letters}, vol. 40, no. 17, pp. 4054--4057, Sep 2015.
	
	\bibitem{Sun16OE}
	Y.~Sun, X.~Yuan, and S.~Pang,
	\newblock ``High-speed compressive range imaging based on active
	illumination,''
	\newblock {\em Opt. Express}, vol. 24, no. 20, pp. 22836--22846, Oct 2016.
	
	\bibitem{Cao16SPM}
	X.~Cao, T.~Yue, X.~Lin, S.~Lin, X.~Yuan, Q.~Dai, L.~Carin, and D.~J. Brady,
	\newblock ``Computational snapshot multispectral cameras: Toward dynamic
	capture of the spectral world,''
	\newblock {\em IEEE Signal Processing Magazine}, vol. 33, no. 5, pp. 95--108,
	Sept 2016.
	
	\bibitem{Yuan16SJ}
	X.~Yuan, H.~Jiang, G.~Huang, and P.~Wilford,
	\newblock ``{SLOPE}: Shrinkage of local overlapping patches estimator for
	lensless compressive imaging,''
	\newblock {\em IEEE Sensors Journal}, vol. 16, no. 22, pp. 8091--8102, November
	2016.
	
	\bibitem{BradyNature12}
	D.~J. Brady, M.~E. Gehm, R.~A. Stack, D.~L. Marks, D.~S. Kittle, D.~R. Golish,
	E.~M. Vera, and S.~D. Feller,
	\newblock ``Multiscale gigapixel photography,''
	\newblock {\em Nature}, , no. 486, pp. 386--389, 2012.
	
	\bibitem{Yuan13ICIP}
	X.~Yuan, J.~Yang, P.~Llull, X.~Liao, G.~Sapiro, D.~J. Brady, and L.~Carin,
	\newblock ``Adaptive temporal compressive sensing for video,''
	\newblock in {\em 2013 IEEE International Conference on Image Processing
		(ICIP)}, Sept 2013, pp. 14--18.
	
	\bibitem{Asif_flatCam}
	M.~S. Asif, A.Ayremlou, A.~Sankaranarayanan, A.~Veeraraghavan, and R.~Baraniuk,
	\newblock ``Flatcam: Thin, bare-sensor cameras using coded aperture and
	computation,''
	\newblock {\em arXiv:1509.00116}, 2015.
	
	\bibitem{cs_Candes06randomProj}
	E.~J. Cand\`{e}s and T.~Tao,
	\newblock ``Near-optimal signal recovery from random projections: universal
	encoding strategies?,''
	\newblock {\em IEEE Transactions on Information Theory}, 2006.
	
	\bibitem{Yuan15Lensless}
	X.~Yuan, H.~Jiang, G.~Huang, and P.~Wilford,
	\newblock ``Lensless compressive imaging,''
	\newblock {\em arXiv:1508.03498}, 2015.
	
	\bibitem{Aharon06TSP}
	M.~Aharon, M.~Elad, and A.~Bruckstein,
	\newblock ``{K-SVD}: An algorithm for designing overcomplete dictionaries for
	sparse representation,''
	\newblock {\em IEEE Transactions on Signal Processing}, vol. 54, no. 11, pp.
	4311--4322, 2006.
	
	\bibitem{Yuan15JSTSP}
	X.~Yuan, T.-H. Tsai, R.~Zhu, P.~Llull, D.~J. Brady, and L.~Carin,
	\newblock ``Compressive hyperspectral imaging with side information,''
	\newblock {\em IEEE Journal of Selected Topics in Signal Processing}, vol. 9,
	no. 6, pp. 964--976, September 2015.
	
	\bibitem{Yuan_16_OE}
	Xin Yuan,
	\newblock ``Compressive dynamic range imaging via bayesian shrinkage dictionary
	learning,''
	\newblock {\em Optical Engineering}, vol. 55, no. 12, pp. 123110, 2016.
	
	\bibitem{Liao14GAP}
	X.~Liao, H.~Li, and L.~Carin,
	\newblock ``Generalized alternating projection for weighted-$\ell_{2,1}$
	minimization with applications to model-based compressive sensing,''
	\newblock {\em SIAM Journal on Imaging Sciences}, vol. 7, no. 2, pp.
	797–--823, 2014.
	
	\bibitem{Yuan16ICIP_GAP}
	X.~Yuan,
	\newblock ``Generalized alternating projection based total variation
	minimization for compressive sensing,''
	\newblock in {\em 2016 IEEE International Conference on Image Processing
		(ICIP)}, Sept 2016, pp. 2539--2543.
	
	\bibitem{Beck09IST}
	A.~Beck and M.~Teboulle,
	\newblock ``A fast iterative shrinkage-thresholding algorithm for linear
	inverse problems,''
	\newblock {\em SIAM J. Img. Sci.}, vol. 2, no. 1, pp. 183--202, Mar. 2009.
	
	\bibitem{Chen10SPT}
	M.~Chen, J.~Silva, J.~Paisley, C.~Wang, D.~Dunson, and L.~Carin,
	\newblock ``Compressive sensing on manifolds using a nonparametric mixture of
	factor analyzers: Algorithm and performance bounds,''
	\newblock {\em IEEE Transactions on Signal Processing}, vol. 58, no. 12, pp.
	6140--6155, December 2010.
	
	\bibitem{Yu12IPT}
	G.~Yu, G.~Sapiro, and St\'ephane Mallat,
	\newblock ``Solving inverse problems with piecewise linear estimators: From
	{G}aussian mixture models to structured sparsity,''
	\newblock {\em IEEE Transactions on Image Processing}, 2012.
	
	\bibitem{Yang14GMM}
	J.~Yang, X.~Yuan, X.~Liao, P.~Llull, G.~Sapiro, D.~J. Brady, and L.~Carin,
	\newblock ``Video compressive sensing using {G}aussian mixture models,''
	\newblock {\em IEEE Transaction on Image Processing}, vol. 23, no. 11, pp.
	4863--4878, November 2014.
	
	\bibitem{Yang14GMMonline}
	J.~Yang, X.~Liao, X.~Yuan, P.~Llull, D.~J. Brady, G.~Sapiro, and L.~Carin,
	\newblock ``Compressive sensing by learning a {G}aussian mixture model from
	measurements,''
	\newblock {\em IEEE Transaction on Image Processing}, vol. 24, no. 1, pp.
	106--119, January 2015.
	
	\bibitem{Yuan15GMM}
	X.~Yuan, H.~Jiang, G.~Huang, and P.~Wilford,
	\newblock ``Compressive sensing via low-rank {Gaussian} mixture models,''
	\newblock {\em arXiv:1508.06901}, 2015.
	
	\bibitem{Kulkarni2016CVPR}
	K.~Kulkarni, S.~Lohit, P.~Turaga, R.~Kerviche, and A.~Ashok,
	\newblock ``Reconnet: Non-iterative reconstruction of images from compressively
	sensed random measurements,''
	\newblock in {\em CVPR}, 2016.
	
\end{thebibliography}

\end{document}